\documentclass[letterpaper]{article} 
\usepackage{aaai24}  
\usepackage{times}  
\usepackage{helvet}  
\usepackage{courier}  
\usepackage[hyphens]{url}  
\usepackage{graphicx} 
\usepackage{natbib}  
\usepackage{caption} 
\usepackage{algorithm}
\usepackage{algorithmic}
\usepackage{amsfonts}
\usepackage{newfloat}
\usepackage{listings}
\usepackage{multirow}

\usepackage{amssymb}
\usepackage{amsmath}

\usepackage[table]{xcolor}
\usepackage{subfig}
\usepackage{booktabs}
\definecolor{baselinecolor}{gray}{.9}
\newcommand{\baseline}[1]{\cellcolor{baselinecolor}{#1}}
\newcommand{\tablestyle}[2]{\setlength{\tabcolsep}{#1}\renewcommand{\arraystretch}{#2}\centering\footnotesize}

\newcommand{\ie}{\textit{i}.\textit{e}.}
\newcommand{\eg}{\textit{e}.\textit{g}.}

\newcommand{\authorskip}{\hspace{2.5mm}}

\usepackage{etoolbox}
\makeatletter
\AfterEndEnvironment{algorithm}{\let\@algcomment\relax}
\AtEndEnvironment{algorithm}{\kern2pt\hrule\relax\vskip3pt\@algcomment}
\let\@algcomment\relax
\newcommand\algcomment[1]{\def\@algcomment{\footnotesize#1}}
\renewcommand\fs@ruled{\def\@fs@cfont{\bfseries}\let\@fs@capt\floatc@ruled
  \def\@fs@pre{\hrule height.8pt depth0pt \kern2pt}%
  \def\@fs@post{}%
  \def\@fs@mid{\kern2pt\hrule\kern2pt}%
  \let\@fs@iftopcapt\iftrue}
\makeatother
\DeclareCaptionStyle{ruled}{labelfont=normalfont,labelsep=colon,strut=off} 
\floatstyle{ruled}
\newfloat{listing}{tb}{lst}{}
\floatname{listing}{Listing}
\title{SupMAE: Supervised Masked Autoencoders Are Efficient Vision Learners}
\author{
    Feng Liang\textsuperscript{\rm 1} \authorskip Yangguang Li\textsuperscript{\rm 2} \authorskip Diana Marculescu\textsuperscript{\rm 1}
}
\affiliations{
    \textsuperscript{\rm 1}The University of Texas at Austin \authorskip
    \textsuperscript{\rm 2}SenseTime Research \\ [2mm]
    
     {\tt\small \{jeffliang, dianam\}@utexas.edu} \qquad {\tt\small liyangguang@sensetime.com}
    
}

     \vspace{-5em}

\usepackage{bibentry}

\begin{document}

\maketitle

\begin{abstract}
Recently, self-supervised Masked Autoencoders (MAE) \cite{mae} have attracted unprecedented attention for their impressive representation learning ability.
However, the pretext task, Masked Image Modeling (MIM), reconstructs the missing local patches, lacking the global understanding of the image.
This paper extends MAE to a \emph{fully-supervised} setting by adding a supervised classification branch, thereby enabling MAE to learn global features from golden labels effectively.
The proposed Supervised MAE (SupMAE) only exploits a visible subset of image patches for classification, unlike the standard supervised pre-training where all image patches are used.
Through experiments, we demonstrate that SupMAE is not only more training efficient, but it also learns more robust features.
Specifically, SupMAE achieves comparable performance with MAE using only 30\% of compute cost when evaluated on ImageNet with the ViT-B/16 model.
SupMAE's robustness on ImageNet variants and transfer learning performance outperforms MAE and standard supervised pre-training counterparts. 
Codes are available at https://github.com/enyac-group/supmae.

\end{abstract}

\section{Introduction}
\label{sec:intro}

Pre-training plays a crucial role in computer vision (CV). 
Supervised pre-training on ImageNet~\cite{imagenet} and then transferring to downstream tasks~\cite{rcnn,fcn,he2017maskrcnn} has revolutionized the entire community.
While it works pretty well for convolutional neural networks (CNNs)~\cite{alexnet,vgg,resnet}, naive ImageNet supervised pre-training does not bring good performance for recently proposed vision transformers (ViT)~\cite{vit}.

\begin{figure}[t]
\begin{center}
\centerline{\includegraphics[width=\columnwidth]{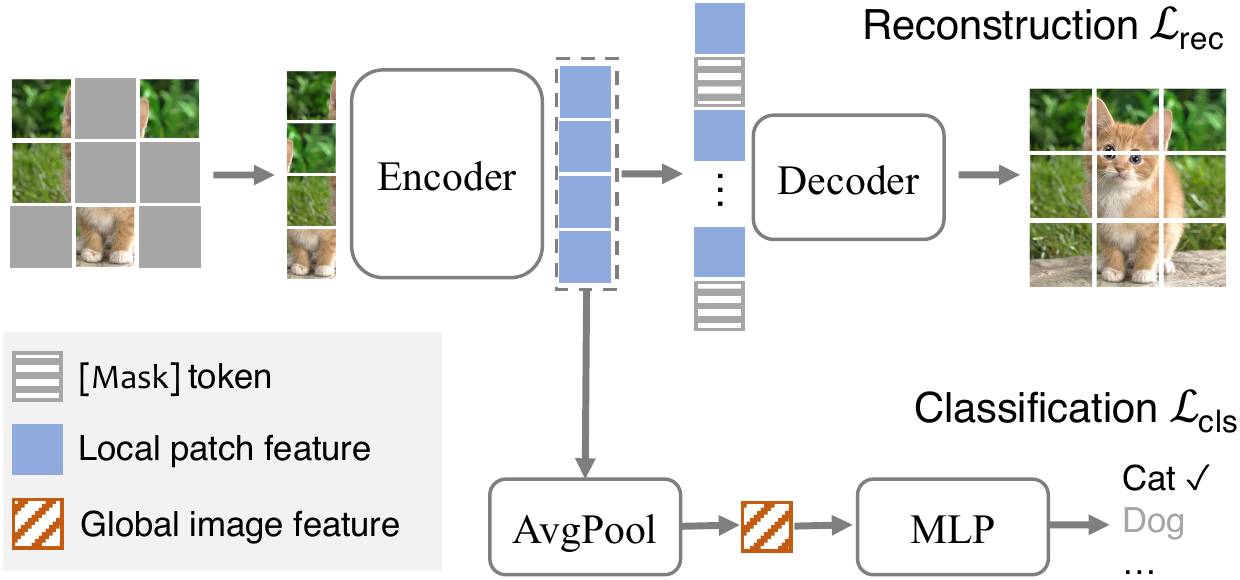}}
\caption{
\textbf{Illustration of the proposed SupMAE method.}
The proposed SupMAE extends MAE by adding a branch for supervised classification in parallel with the existing reconstruction objective. 
In the pre-training phase, only a subset of the visible patches is processed by a ViT encoder.
Their corresponding patch features are used to (1) reconstruct the missing pixels and (2) classify the category.
In the fine-tuning phase, the encoder is applied to uncorrupted images for recognition tasks.}
\label{fig:supmae_overview}
\end{center}
\vskip -0.2in
\end{figure}

To unearth the potential of ViT, self-supervised pre-training methods~\cite{mocov3,dino,beit,mae,xie2021simmim} are emerging as an alternative. 
Among them, the Masked Autoencoder (MAE)~\cite{mae} is the state-of-the-art method that adopts a BERT-type masked autoencoding scheme~\cite{bert}.
As shown in the top part of Figure~\ref{fig:supmae_overview}, MAE masks random patches of the input image and reconstructs the missing pixels with the visible patches.
Although it achieves remarkable performance, MAE requires thousands of epochs to pre-train.
This is because the key ingredient of MAE, Masked Image Modeling (MIM), only learns better middle-level interactions among patches~\cite{a2mim}.
No global features, \ie, features that can represent the entire image, are learned during pre-training.

How can we incorporate global feature learning into MAE?
A natural solution is to leverage the golden labels, which allows MAE to know what concept it is reconstructing.
However, whether the use of supervised labels would benefit MAE has not been explored yet.
~\citet{mae} show that standard supervised pre-training underperforms even with more data augmentations~\cite{deit} or stronger regularization~\cite{howtotrainvit}.

In this paper, we first show that supervised pre-training can benefit MAE in training efficiency, model robustness, and transfer learning ability. 
The proposed method, Supervised MAE (SupMAE), extends MAE by adding a branch for supervised classification in parallel with the existing reconstruction objective (depicted in Figure~\ref{fig:supmae_overview}).
Formally, the subset of visible patches is fed into the ViT encoder to obtain local patch features.
For the reconstruction branch, patch features and padded \texttt{mask} tokens are processed by a small decoder that reconstructs the original image.
For the classification branch, an average pooling operator is applied on patch features to get the global image feature which an MLP follows for classification.
During fine-tuning, we only use the ViT encoder for downstream recognition tasks.

Unlike standard supervised pre-training methods~\cite{deit,howtotrainvit} that use all patch features (or the equivalent \texttt{class} token), SupMAE only uses a subset of the visible patches to do classification.
This is mainly based on the intuition that images are highly spatially redundant: humans can easily recognize an image even with a partial observation (a subset of patches).
This design also makes SupMAE more sample-efficient: we can compute loss over \emph{all} input tokens during training rather than just the subset that is masked out\footnote{MAE only computes the loss on masked patches.}. 
From the perspective of data augmentation, random masking can generate different training samples for each iteration, serving as a strong regularization during supervised pre-training.

Through empirical experiments, we demonstrate that: 
(1). SupMAE is training efficient. 
SupMAE ViT-B/16 achieves 83.6\% accuracy when fine-tuned on ImageNet-1K using only 400 pre-training epochs, 3$\times$ less compared with 1600 epochs of MAE. 
We further benchmark the wall-clock running time of different supervised and self-supervised pre-training methods on GPU, showing the effectiveness and efficiency of the proposed SupMAE. 
(2). SupMAE is more robust to natural corruption. 
Notably, our SupMAE outperforms the MAE counterpart by an average gain of 1.8\% on the robustness benchmark containing four ImageNet variants.
(3). SupMAE learns more transferable features. 
Few-shot linear probing and fine-tuning on 20 downstream classification datasets show the superior performance of SupMAE. 
Moreover, our fine-tuned ViT-B/16 model achieves 49.0\% mIoU on the ADE20K semantic segmentation validation dataset, revealing that the supervised pre-training can benefit the dense downstream tasks as well.

To summarize, the contributions of our paper are three-fold:

\begin{itemize}
    \item To the best of our knowledge, this is the first work to study whether supervised pre-training can benefit MAE. This is a direction that intuitively makes sense because golden labels allow MAE to know what concept it is reconstructing.
    \item The proposed SupMAE only uses a subset of the visible patches to do classification rather than standard supervised pre-training that uses all patches. This design also makes SupMAE more sample-efficient: we can utilize \emph{all} input tokens during training rather than just the subset that is masked out. 
    \item Through empirical experiments, we demonstrate SupMAE is more training efficient and it also learns more robust features and transferable features.
\end{itemize}

\section{Related Work}
\label{sec:related_work}

\textbf{Supervised vision transformers} 
Transformers~\cite{transformer}, a self-attention-based architecture that originated in Natural Language Processing (NLP), has emerged as an alternative to CNNs in CV. 
Due to the lack of inductive biases inherent to CNNs, the first Vision Transformers (ViT) proposed in ~\citet{vit} do not perform well when trained on the mid-scale dataset ImageNet.
Researchers have proposed to design data augmentations~\cite{deit,touvron2022deit3} or add more regularization~\cite{howtotrainvit} to improve the generality of supervised ViT.
The proposed SupMAE is a supervised pre-training method built upon the recently introduced MAE~\cite{mae}.
Compared with the aforementioned supervised methods, SupMAE only uses a \emph{subset of the visible patches} to do classification rather than all patches.

\textbf{Masked image modeling (MIM)} 
Generative pre-training, \eg, BERT~\cite{bert} and GPT~\cite{GPT}, has been a well-established paradigm in NLP.
Recently, researchers have been trying to introduce generative pre-training into CV.
The pioneering work iGPT~\cite{igpt} adopts GPT-type auto-regressive modeling that predicts following pixels auto-regressively.
More work adopts BERT-type Masked Image modeling (MIM) as the pretext task: representations are learned through reconstructing the missing part in the image.
The reconstruction objective can be raw pixels~\cite{mae,xie2021simmim}, discrete visual tokens~\cite{beit,peco}, low-level local features~\cite{wei2021masked}, or latent representations~\cite{baevski2022data2vec}.
The proposed SupMAE is built upon MAE, the representative of generative approaches.

\textbf{Contrastive learning} 
Another line of work for self-supervised visual representation learning is contrastive learning~\cite{mocov3,dino,moby,assran2022msn}
SupCon~\cite{supcon} is most relevant to our work: it extends contrastive learning into the supervised setting while we extend MIM into the supervised setting.

\textbf{Multi-objective pre-training} 
Our work is also highly related to multi-objective pre-training, where the model is trained with multiple auxiliary tasks.
~\citet{zhang2016augmenting,le2018supervised} augments the supervised CNNs with unsupervised reconstruction objective.
Our SupMAE shares the same spirit but differs from these classical methods in numerous ways.
More recently, Repre~\cite{repre} incorporates local feature learning into contrastive methods via reconstructive pre-training.
CMAE~\cite{cmae} combines contrastive learning with MAE to further strengthen the representation.
Different from them, SupMAE extends the MAE into a fully supervised setting to enable MAE to effectively learn global features from golden labels.

\section{SupMAE Method}
\label{sec:method}

The overall framework of the proposed SupMAE is illustrated in Figure~\ref{fig:supmae_overview}.
SupMAE consists of three components: encoder, reconstruction decoder, and classification head.
There are two complementary objectives: the reconstruction objective and the supervised classification objective.
Details are introduced as follows.

\begin{algorithm}[th!]
\caption{SupMAE: PyTorch-like pseudo code}
\label{alg:code}
\algcomment{
\textbf{Notes}: 
\texttt{class} and \texttt{mask} tokens as well as positional embeddings are omitted for simplicity. For more details, please refer to our whole code in the supplementary files. 
}
\definecolor{codeblue}{rgb}{0.25,0.5,0.5}
\definecolor{codekw}{rgb}{0.85, 0.18, 0.50}
\begin{lstlisting}[language=python,numbers=none,basicstyle=\footnotesize\ttfamily,mathescape]
# f_enc / f_dec: encoder / decoder
# f_cls: mlp classification head
# mask_r: mask ratio
# $\color{codeblue}\lambda_{rec}$, $\color{codeblue}\lambda_{cls}$: objective weights
# tau: temperature

for x, tgt in loader:  # load a minibatch
    x = patch_emb(x) # embed patches
    x_v, x_m = masking(x, mask_r) # ramdom split visible and masked patches
    q_v = f_enc(x_v) # local patch features
    logits = f_cls(q_v) # predicted labels
    k_m = f_dec(q_v) # reconstructed pixels
    loss = $\lambda_{rec}$ * rec_loss(k_m, x_m) + $\lambda_{cls}$ * cls_loss(logits, tgt)
    loss.backward()
    update()  # optimizer update

def rec_loss(k, x): # reconstruction 
    x = norm_pix(x) # normalize every patch
    loss = (k - x) ** 2 # compute MSE loss over masked patches
    return loss

def cls_loss(logits, tgt): # classification
    loss = CrossEntropyLoss(logits/tau, tgt)
    return loss
    
\end{lstlisting}
\end{algorithm}

\textbf{Image masking and encoding} 
Following ~\citet{vit}, the input image $\mathbf{x} \in \mathbb{R}^{H \times W \times C}$ is first divided into several non-overlapping patches $\mathbf{x}_p \in \mathbb{R}^{N \times (P^2 \cdot C)}$, where $(H, W, C)$ corresponds to the resolution and channels of the original image, $(P, P)$ is the resolution of each patch and $N=HW/P^2$ is the number of non-overlapping patches.
The patches are fed into a linear projection (a.k.a., PatchEmbed) to get patch embeddings.
Following ~\citet{mae}, we mask a large portion of patches (\eg, 75\%).
We denote the visible patches and masked patches as $\mathbf{x}^{v}$ and $\mathbf{x}^{m}$, respectively.
The remaining visible patches $\mathbf{x}^{v}$ are added to positional embeddings and then processed by a ViT encoder~\cite{vit} to get corresponding local patch features $\mathbf{q}^{v}$.
The subset of the visible patch features $\mathbf{q}^{v}$ is used for reconstruction and classification, introduced next.

\textbf{Reconstruction branch} 
Due to the random masking, the length of the visible patch features $\mathbf{q}^{v}$ is shorter than the image patch length $N$.
Thus, we pad patch features $\mathbf{q}^{v}$ with \texttt{mask} tokens~\cite{bert} to generate a full set of features. 
Each \texttt{mask} token is a shared, learned vector that indicates the presence of a missing patch to be reconstructed. 
As in the encoding, this full set of features is added with positional embeddings and processed by a transformer-based decoder. 
We find that the decoder of SupMAE can be very light-weight, \eg, a one-layer transformer, which is consistent with the findings in MAE. 
After the decoding, a linear layer (omitted in Figure~\ref{fig:supmae_overview}) projects features into the pixel space.
The reconstruction objective $\mathcal{L}_{rec}$, mean squared error (MSE) in our SupMAE, is operated between the reconstructed and original images. 
Following prior work~\cite{mae,bert}, we compute the loss only on masked patches.

\textbf{Classification branch} 
The same set of visible patch features $\mathbf{q}^{v}$ are further used for supervised classification.
This is different from standard supervised pre-training, where all patches are used. 
More formally, a global pooling first condenses local patch features into the global representation of the image.
The global representation is then used to predict the golden labels.
The classification branch is complementary to the reconstruction branch from two perspectives: (1) the classification branch brings global feature learning into the framework, (2) we can compute loss over \emph{all} input tokens during training rather than just the subset that is masked out (where reconstruction operates on).

For the classification head, we use a two-layer MLP, with Batchnorm~\cite{ioffe2015batchnorm} and ReLU activation function injected in between.
We further ablate the number of linear layers in Table~\ref{tab:mlp_layers}.
We find it is also beneficial to introduce temperature $\tau$~\cite{hinton2015distilling} as a parameter after the prediction of classification head (a.k.a. logits).
The temperature $\tau$ is a parameter that controls the concentration level of the distribution, which is widely used in supervised~\cite{wang2017normface}/self-supervised~\cite{wu2018unsupervised,mocov3} feature learning.
$\tau$ is set to 10 in our experiments.
The classification loss $\mathcal{L}_{rec}$, cross-entropy (CE) in our SupMAE, is performed between predicted and golden labels.
An alternative to representing the entire image is the \texttt{class} token, which is left 'untouched' during pre-training. Through empirical study, we find that global pooling brings better results than the \texttt{class} token (see Table~\ref{tab:cls_token}).

\textbf{Overall objective} Our SupMAE is optimized with reconstruction loss and classification loss, which simultaneously learns fine-grained local and global features. 
We use a weighted sum of these two loss functions as our overall loss as follows:

\vspace{-1em}
\begin{equation}
\mathcal L = \lambda_{rec}\mathcal L_{rec} + \lambda_{cls}\mathcal L_{cls}
\label{eq:rec_cls_loss}
\end{equation}
where $\lambda_{rec}$, $\lambda_{cls}$ are weights to balance the two objectives. $\lambda_{rec}$, $\lambda_{cls}$ are set as 1.0 and 0.01, respectively (see Table~\ref{tab:cls_ratio}). 

We further show the Pytorch-like pseudo-code in Algorithm~\ref{alg:code}. Please refer to our whole code in the supplementary files for more details.

\textbf{Fine-tuning on downstream tasks} 
After pre-training, our SupMAE model can be further fine-tuned on target recognition tasks to achieve better performance.
During fine-tuning, we only keep the encoder while discarding the decoder and classification head.
We use the full set of patches, \ie, uncorrupted images when fine-tuning downstream recognition tasks.


\begin{table*}[t]
\centering
\caption{
\textbf{Comparison with supervised and self-supervised pre-training methods} All methods are using ViT-B/16 model. 
Besides the number of pre-training (PT) and fine-tuning (FT) epoch, we further estimate the wall-clock time for PT and FT, benchmarked on 8 A5000 GPUs. The normalized cost is relative to SupMAE. SupMAE shows a great efficiency and can achieve the same accuracy as MAE using only 30\% compute.
}
\label{tab:main_results}
\vspace{-0.5em}
\small
\begin{tabular}{lccccccc}
\toprule
method    & \multicolumn{1}{c}{PT epochs} & \begin{tabular}[c]{@{}c@{}}PT cost\\ (Hours)\end{tabular} & \multicolumn{1}{c}{FT epochs} & \begin{tabular}[c]{@{}c@{}}FT cost\\ (Hours)\end{tabular} & \multicolumn{1}{c}{\begin{tabular}[c]{@{}c@{}}Total cost\\ (Hours)\end{tabular}} &  \begin{tabular}[c]{@{}c@{}}Normalized\\ cost\end{tabular} & \begin{tabular}[c]{@{}c@{}}Top1\\ acc.\end{tabular} \\
\midrule
\multicolumn{8}{c}{\textit{Self-Supervised pre-training methods.}} \\
MoCov3~\cite{mocov3}    &  300      &  250           &  150      &   45.7          &   295.7     &  2.35$\times$   &   83.2     \\
BEiT~\cite{beit}    &  800     &   233.3        &   100     &   31.5     &  264.8    &  2.10$\times$   &   83.2     \\
MAE~\cite{mae}    &  1600      &  364           &  100      &   30          &    394    & 3.12$\times$    &  \textbf{83.6}      \\
\midrule
\multicolumn{8}{c}{\textit{Supervised pre-training methods.}} \\
ViT~\cite{vit}     &  -  &    -        & -       &     -       &   -    & -     &   77.9    \\
DeiT~\cite{deit}     &  300  &    91.5         & -       &     -       &   91.5    & 0.73$\times$     &   81.8     \\
Naive supervised~\cite{mae} &  300      &    90         &  -      & -         &   90    &  0.71$\times$    &   82.3     \\
SupMAE(Ours)    &  400      &    95.9         &  100   &  30           &  125.9      &  1$\times$   &    \textbf{83.6}                  \\
\bottomrule
\end{tabular}
\end{table*}

\section{Experimental Results}
\label{sec:experiment}

\subsection{Main results on ImageNet-1K}

We first conduct our experiments on widely used ImageNet-1K dataset~\cite{imagenet}. 
We basically follow MAE ~\cite{mae} for the setup and training hyper-parameters. 
Details are in our supplementary files.

\textbf{Comparison with other pre-training methods} 
We compare our SupMAE with other supervised or self-supervised pre-training methods in Table~\ref{tab:main_results}. 
All methods use the same ViT-B/16 architecture.
For MAE and our SupMAE, the decoder is a one-layer transformer, and the mask ratio is set to 75\%.
The number of epochs is most widely used as an indicator of the training cost. 
However, it might be misleading as different methods require different time to run one epoch.
Thus, we further benchmark the real wall-clock time of pre-training (PT) and fine-tuning (FT) on 8 NVIDIA A5000 24 GB GPU server with Pytorch.
We use the official repository of Mocov3, DeiT, BEiT, and MAE to estimate the cost.
Interestingly, we observe that the one-epoch cost of different methods varies greatly.
The one-epoch cost of SupMAE is only $\sim$30\% of MoCov3~\cite{mocov3} because SupMAE's encoder only processes a small subset of tokens.
We also observe that data loading time is relatively negligible; most of the cost is computing.

We compare SupMAE with three supervised counterparts: ViT~\cite{vit}, DeiT~\cite{deit} and naive supervised results from ~\citet{mae}. 
SupMAE can achieve 83.6\% ImageNet top1 accuracy, significantly outperforming the standard supervised methods.
When compared with the self-supervised method, such as MoCo-v3~\cite{mocov3}, BEiT~\cite{beit} and MAE~\cite{mae}, SupMAE can achieve comparable results with much lower training compute.
This shows that supervised pre-training is compatible with self-supervised objectives and can improve the training efficiency of self-supervised methods.
The efficiency and effectiveness of SupMAE lie on two facts: (1) the reconstruction objective helps to learn better local features (2) the classification objective provides the global feature learning ability.

\begin{figure}[t]

\begin{center}
\centerline{\includegraphics[width=1.0\columnwidth]{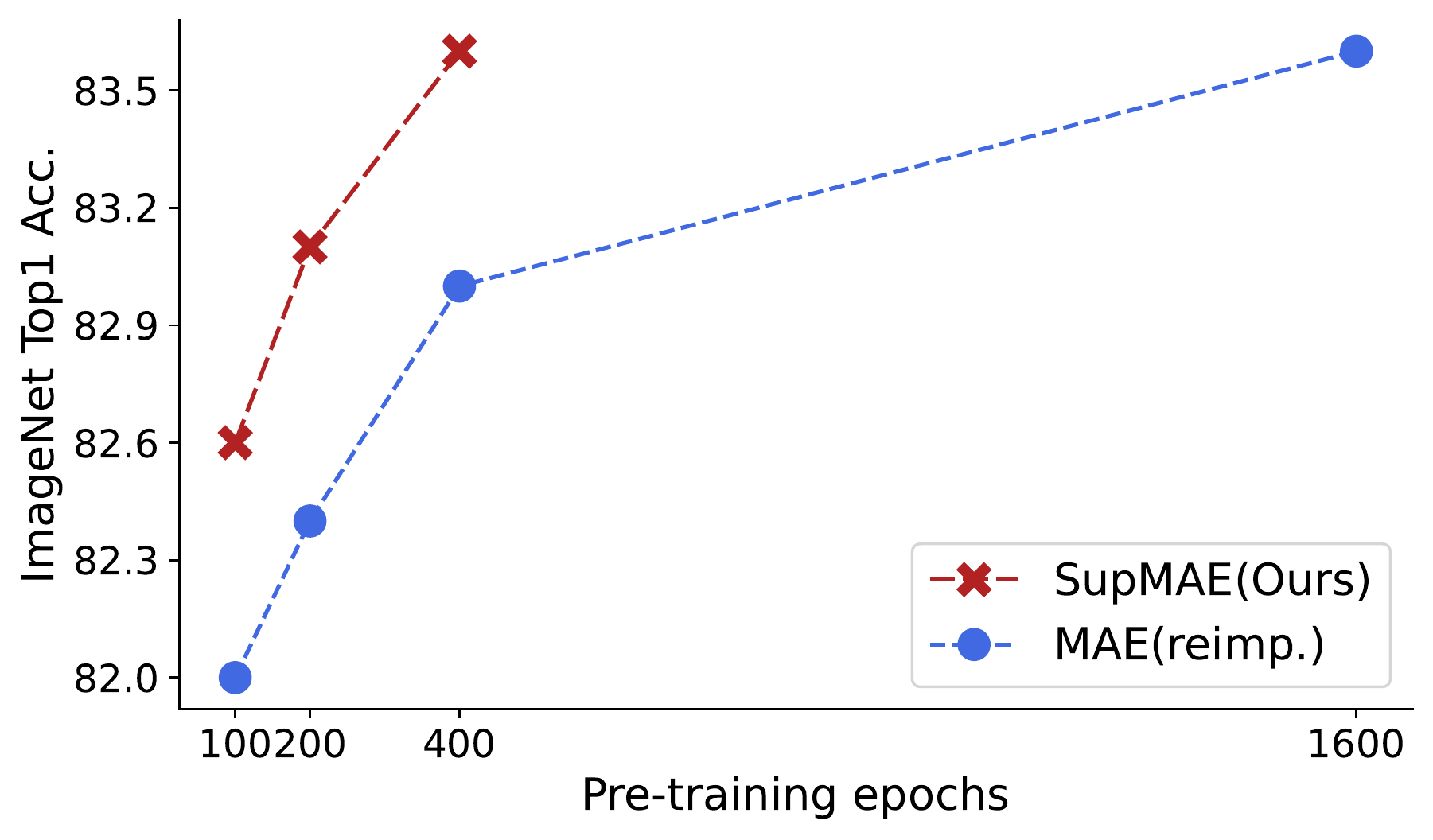}}
\vspace{-1em}
\caption{\textbf{Performance of different pre-training epochs}
Comparison between MAE and SupMAE when pre-trained for different epochs. SupMAE is efficient and shows a much faster convergence speed.}
\label{fig:different_epochs}
\vspace{-1em}
\end{center}
\vspace{-2em}
\end{figure}

\textbf{Different pre-training epochs} 
We compare different pre-training epochs of MAE and our proposed SupMAE in Figure~\ref{fig:different_epochs}. 
The encoder is ViT-B/16 architecture, and the mask ratio is set to 75\%.
We use a one-layer transformer decoder for SupMAE, unlike an 8-layer transformer decoder for MAE~\cite{mae}.
We only change the number of pre-training epochs, \emph{i.e.}, all models are fine-tuned for 100 epochs.
When pre-trained for very few epochs, such as 100 epochs, SupMAE can achieve 82.6\% accuracy, which is 0.6\% higher than the MAE counterpart.
When pre-trained for 200 epochs, SupMAE achieves 83.1\% accuracy, higher than directly training from scratch 82.3\% with 14\% less time cost (78h \textit{vs}. Naive supervised's 90h).
Our SupMAE can achieve 83.6\% accuracy with only 400 pre-training epochs, while the MAE counterpart needs 1600 epochs to achieve such performance.

Unfortunately, we do not observe further improvement when we pre-train our SupMAE longer, such as for 800 epochs. 
Pre-training 800 epochs yields almost the same accuracy as 400 epochs.
We conjecture there might be two reasons: 
(1) SupMAE can be categorized into supervised pre-training. However, supervised pre-training is easy to saturate, \emph{e.g.}, DeiT~\cite{deit} saturates at 300 epochs even with a strong augmentation.
The proposed SupMAE is also suffering from this problem.
(2) The fine-tuning recipe is relatively mature, and we might already exploit most of the information in ImageNet with the ViT-B/16 model. 
Since we follow almost all hyperparameters of MAE, we conjecture more dedicated hyperparameter tuning might lead to better results.

\subsection{Robustness evaluation on ImageNet variants}

In this section, we compare the robustness performance of MAE~\cite{mae}, DeiT~\cite{deit} and our SupMAE on four ImageNet variants.
These datasets compare the robustness from different perspectives, including (1) IN-Corruption~\cite{imagenet-corruption} for common corruptions (2) IN-Rendition~\cite{imagenet-rendition} for semantic shifts (3) IN-Adversarial~\cite{imagenet-adversarial} for natural adversarial examples and (4) IN-Sketch~\cite{imagenet-sketch} for color and texture shifts.

We test the same SupMAE model as in Table~\ref{tab:main_results} on these ImageNet variants without any specialized fine-tuning. The results are summarized in Table~\ref{tab:robustness}. Our SupMAE significantly outperforms MAE in three out of four datasets. To further validate the average robustness, we further benchmark the robustness score, which is an average score across four datasets. Since the IN-Corruption uses mean corruption error, we use '100 - error' as its metric when averaging the score. As we can see, SupMAE achieves a 43.6\% average score, which significantly outperforms self-supervised MAE (+1.8\%) and supervised DeiT (+4.2\%).

\begin{table}[t]
\caption{
\textbf{Robustness evaluation on robustness benchmark.} 
All methods use the same ViT-B/16 architecture.
The metric is top-1 accuracy, except for IN-Corruption~\cite{imagenet-corruption} which uses mean corruption error.
We test the same SupMAE model as in Tabel~\ref{tab:main_results} on 4 ImageNet variants \emph{without} any specialized fine-tuning. The score is measured by the averaging metric across four variants (we use '100 - error' for the IN-Corruption performance metric).
DeiT results are reproduced using the official checkpoint.
Our SupMAE model shows better robustness on the benchmark.
}
\label{tab:robustness}
\small
\centering
\begin{tabular}{lccc}
\toprule
dataset        & MAE           & DeiT          & SupMAE(Ours)  \\
\midrule
IN-Corruption $\downarrow$  & 51.7          & \textbf{47.4} & 48.1          \\
IN-Adversarial & \textbf{35.9} & 27.9          & 35.5          \\
IN-Rendition   & 48.3          & 45.3          & \textbf{51.0} \\
IN-Sketch      & 34.5          & 32.0          & \textbf{36.0} \\
Score         & 41.8          & 39.5          & \textbf{43.6} \\
\bottomrule
\end{tabular}
\end{table}

\subsection{Transfer Learning experiments}
\label{exp:few_shot}

\subsubsection{Few-shot learning on 20 classification datasets}
We adopt the ELEVATOR~\cite{elevater} benchmark to conduct few-shot transfer learning on 20 image classification datasets. Table~\ref{tab:few-shot} reports the averaged accuracy of three methods: MAE, MoCo-v3, and the proposed SupMAE. 
More details about the datasets and breakdown results can be found in supplementary files.
For a fair comparison, all the models use the same automatic hyper-parameter tuning process as in ~\citet{elevater}, and no model-/dataset-specific hyper-parameter tuning is employed.
The average accuracy can represent the model transferability under a few-shot setting.

We observe that MAE performs worst under both linear probing and fine-tuning scenarios. We conjecture that this is caused by the fact that MAE can learn good local features but lacks the global image understanding. 
Our SupMAE performs significantly better than its MAE counterpart, thanks to the introduced global feature learning during pre-training.
Under an extreme 5-shot linear probing setting, SupMAE outperforms its MAE counterpart by a large +14.6\% margin.
Our SupMAE shows better performance when we fine-tune the model end-to-end.
We conjecture that this is because SupMAE only observe some fragments of images, \eg, 25\%, during the pre-training.
End-to-end fine-tuning can adapt the model weights to suit the complete images better.
Notably, SupMAE can outperform MoCo-v3 by a considerable +1.4\% margin under 50-shot fine-tuning setting.

\begin{table}[t!]
    \centering
    \footnotesize
    \caption{\textbf{Few-shot transfer learning.} 
    All methods use the same ViT-B/16 architecture.
    We report the linear probing and fine-tuning averaged scores on 20 image classification datasets.
    X-shot denotes the number of labeled images per category used during transfer learning. 
    Our SupMAE significantly outperforms its MAE counterpart.
    MAE and MoCo-v3 results are from ~\citet{elevater}.
    }
    \label{tab:few-shot}
    \scalebox{0.92}{
\begin{tabular}{@{}p{2.2cm}@{} c | ccc}
    \toprule
       \multicolumn{2}{c|}{ \multirow{1}{*}{Pre-training Settings} }  &  \multicolumn{3}{c}{ 20 Image Classification Datasets }  \\
       \cmidrule{3-5} 
       Checkpoint  & Method &  5-shot & 20-shot & 50-shot  \\
        \midrule
         \multicolumn{5}{c}{ \bf{Linear Probing} } \\
        MAE & Self-Sup. & 33.37 {\tiny $\pm$ 1.98} & 48.03 {\tiny $\pm$ 2.70} & 58.26 {\tiny $\pm$ 0.84} \\  
        MoCo-v3 & Self-Sup. & \textbf{50.17 {\tiny $\pm$ 3.43}} & \textbf{61.99 {\tiny $\pm$ 2.51}} & \textbf{69.71 {\tiny $\pm$ 1.03}} \\
        SupMAE(Ours) & Sup.  & 47.97 {\tiny $\pm$ 0.44} & 60.86 {\tiny $\pm$ 0.31} & 66.68 {\tiny $\pm$ 0.47} \\ 
        \midrule
          \multicolumn{5}{c}{ \bf{Fine-tuning } } \\
        MAE & Self-Sup.  & 36.10 {\tiny $\pm$ 3.25} & 54.13 {\tiny $\pm$ 3.86} & 65.86 {\tiny $\pm$ 2.42} \\ 
        MoCo-v3  & Self-Sup. & 39.30 {\tiny $\pm$ 3.84} & 58.75 {\tiny $\pm$ 5.55} & 70.33 {\tiny $\pm$ 1.64}  \\
        SupMAE(Ours) & Sup.  & \textbf{46.76 {\tiny $\pm$ 0.12}} & \textbf{64.61 {\tiny $\pm$ 0.82}} & \textbf{71.71 {\tiny $\pm$ 0.66}} \\ 
        \bottomrule
    \end{tabular}
    }
\end{table}

\subsubsection{Semantic segmentation in ADE20k}

\begin{table}[t]
\small
\centering
\caption{\textbf{Transferring to semantic segmentation on ADE20K} 
All methods use UperNet with ViT-B/16 backbone. 
For a fair comparison with supervised methods, we use a fine-tuned model for MAE and SupMAE.
Naive supervised results are from ~\citet{mae}.
MAE results are reproduced using the official fine-tuned checkpoint.
}
\label{tab:ade20k}
\begin{tabular}{llcc}
\toprule
method & mIoU & aAcc & mAcc \\
\midrule
Naive supervised & 47.4 & - & - \\
MAE & 48.6 & \textbf{82.8} & 59.4 \\
SupMAE (ours) & \textbf{49.0} & 82.7 & \textbf{60.2} \\
\bottomrule
\end{tabular}
\end{table}

\begin{table*}[t]
\caption{\textbf{SupMAE ablation experiments} 
All experiments are using ViT-B/16 on ImageNet-1K. We report fine-tuning (ft) and linear probing (lin) accuracy (\%). If not specified, the default is: the loss ratios of reconstruction (rec) and classification (cls) are 1 and 0.01, global pooling feature is used for classification, the decoder has depth 8, the data augmentation is random resized cropping, the masking ratio is 75\%, and the pre-training length is 200 epochs. Default settings are marked in \colorbox{baselinecolor}{gray}.
}
\label{tab:ablations} 
\small
\vspace{-.5em}
\centering
\subfloat[
\textbf{Pre-training objectives}. Reconstruction and classfication supports each other.  \\
\label{tab:rec_cls_obj}]{
\centering
\begin{minipage}{0.3\linewidth}
\vspace{-1.2em}
{\begin{center}
\begin{tabular}{cccc}
\toprule
rec & cls & ft & lin \\
\midrule
\checkmark & & 82.4 & 58.0 \\
& \checkmark & 79.9 & 59.9 \\
\checkmark & \checkmark & \baseline{\textbf{83.1}} & \baseline{\textbf{70.1}} \\
\bottomrule
\end{tabular}
\end{center}}
\end{minipage}
}
\hspace{2em}
\subfloat[
\textbf{\texttt{Class} token}. Global pooling feature performs better than the additional \texttt{class} token.
\label{tab:cls_token}]{
\centering
\begin{minipage}{0.29\linewidth}
{\begin{center}
\begin{tabular}{ccc}
\toprule
case & ft & lin \\
\midrule
cls token & 79.1 & 65.8 \\
global pool & \baseline{\textbf{83.1}} & \baseline{\textbf{70.1}} \\
\bottomrule
\end{tabular}
\end{center}}
\end{minipage}
}
\hspace{2em}
\subfloat[
\textbf{Data augmentation}. Our SupMAE works with minimal data augmentation like MAE.
\label{tab:data_aug}
]{
\centering
\begin{minipage}{0.29\linewidth}
{\begin{center}
\begin{tabular}{ccc}
\toprule
data aug & ft & lin \\
\midrule
randcrop & \baseline{\textbf{83.1}} & \baseline{70.1} \\
randcrop,cjit & 83.0 & \textbf{70.3} \\
\bottomrule
\end{tabular}
\end{center}}
\end{minipage}
}
\\
\centering
\subfloat[
\textbf{Loss ratio}. Small classification loss ratio works best.  \\
\label{tab:cls_ratio}]{
\centering
\begin{minipage}{0.3\linewidth}
{\begin{center}
\vspace{-1.2em}
\begin{tabular}{ccc}
\toprule
cls ratio & ft  & lin  \\
\midrule
0.02 & 82.9 & \textbf{70.2} \\
0.01 & \baseline{\textbf{83.1}} & \baseline{70.1} \\
0.005 & \textbf{83.1} & 69.8 \\
0.002 & 82.8 & 68.8 \\
\bottomrule
\end{tabular}
\end{center}}
\end{minipage}
}
\hspace{2em}
\subfloat[
\textbf{Decoder depth}. SupMAE works well with a light decoder, 
\ie, an one-layer transformer decoder.
\label{tab:decoder}
]{
\centering
\begin{minipage}{0.29\linewidth}
{\begin{center}
\begin{tabular}{ccc}
\toprule
blocks & ft & lin \\
\midrule
1 & \textbf{83.1} & 65.7 \\
4 & \textbf{83.1} & 68.2 \\
8 & \baseline{\textbf{83.1}} & \baseline{\textbf{70.1}}  \\
\bottomrule
\end{tabular}
\end{center}}
\end{minipage}
}
\hspace{2em}
\subfloat[
\textbf{MLP layers}. An appropriate number of layers should be set for the classification head. 
\label{tab:mlp_layers}
]{
\centering
\begin{minipage}{0.29\linewidth}
{\begin{center}
\begin{tabular}{ccc}
\toprule
mlp layers & ft & lin \\
\midrule
1 & 83.0 & \textbf{72.5} \\
2 & \baseline{\textbf{83.1}} & \baseline{70.1} \\
3 & 82.9 & 69.5  \\
\bottomrule
\end{tabular}
\end{center}}
\end{minipage}
}
\vspace{-2em}
\end{table*}

We adopt the UperNet~\cite{upernet} as the semantic segmentation model on challenging ADE-20k~\cite{ade20k} dataset.
For a fair comparison, we follow the same training configuration and code as MAE~\cite{mae} with the MMSegmentation~\cite{mmseg2020} framework. 
The models are trained on 8 GPUs with a total batch size of 16 for 160k iterations.
The input resolution is set to 512$\times$512.

We compare three methods in Table~\ref{tab:ade20k}: naive supervised, MAE, and the proposed SupMAE. Naive supervised indicates the supervised pre-training done from scratch, in which we directly use the reported mIoU from ~\citet{mae}.
We initialized the segmentation models using model weights after supervised fine-tuning on ImageNet-1K for two reasons: (1) it is a fair comparison with supervised methods (2) supervised fine-tuned weights can bring superior performance than the self-supervised pre-trained weights.
Our SupMAE outperforms its MAE counterpart by a considerable +0.4\% margin, showing that the supervised pre-training
can benefit the dense downstream tasks as well

\subsection{Ablation studies}
\label{sec:ablation}
To verify the components in the proposed SupMAE, we further conduct ablation studies using the default settings in Table ~\ref{tab:ablations} (see caption).

\textbf{Pre-training objectives.} Table~\ref{tab:rec_cls_obj} studies the pre-training objectives. 
As depicted in Figure~\ref{fig:supmae_overview}, the method degrades into MAE with only the reconstruction (rec) objective.
If we only use the classification (cls) objective, the method degrades into standard supervised pre-training (with 75\% input patches masked out).
We find that neither the reconstruction nor the classification objective can perform well when used in isolation. 
This is because only with both objectives can 100\% patches be exploited: 
(1) reconstruction operates on 75\% masked patches, and (2) classification operates on 25\% visible patches.

\textbf{Class token.} Table~\ref{tab:cls_token} shows the impact of using the \texttt{class} token during pre-training.
As introduced in ViT~\cite{vit}, the additional \texttt{class} token is broadly viewed as the representation of the entire image. 
However, we find it better to use the global pooling feature (pooling over patch features) rather than \texttt{class} token in SupMAE.
We conjecture that the \texttt{class} token may not work well when only operating on a subset of visible patches.

\textbf{Data augmentation.} Table~\ref{tab:data_aug} shows the influence of the data augmentation.
Unlike other supervised pre-training methods, such as DeiT, which use a very heavy data augmentation, we find our SupMAE works pretty well with a minimal augmentation, \emph{e.g.}, random resized cropping (randcrop).
Adding color jittering (cjit) does not bring further improvements.
This finding is consistent with MAE.
This is because random masking already generates different training samples for each iteration.
Thus, less data augmentation is required to regularize training.

\textbf{Loss ratio.} Table~\ref{tab:cls_ratio} studies different ratios of classification (cls) loss. 
To balance two pre-training objectives (see Equation~\ref{eq:rec_cls_loss}), we first fix the ratio of reconstruction loss to 1, and then we tune the classification ratio.
We find that a small classification loss ratio works best.
Too large a ratio forces SupMAE to degrade into standard supervised pre-training, which hurts performance (see Table~\ref{tab:rec_cls_obj}).

\textbf{Decoder depth.} Table~\ref{tab:decoder} studies different depths of the decoder. 
As we increase the depth of the decoder, the linear probing (lin) accuracy increases steadily, but the end-to-end fine-tuning (ft) accuracy stays the same.
The finding is also consistent with MAE.
This is due to the gap between pixel reconstruction and classification: a shallow decoder would require the output of the encoder to be more specialized for reconstruction, which is harmful to linear probing.

\textbf{MLP layers.} Table~\ref{tab:mlp_layers} shows the influence of the number of MLP layers in the classification head.
Increasing the layers decreases the linear probing (lin) accuracy.
We conjecture that this is because one linear layer makes the encoder output linearly separable, which is suitable for linear probing.
However, since SupMAE has a reconstruction objective, we may want the latent representations at a more abstract level.
Thus, we choose a two-layer MLP as default setting.

\subsection{Discussion}

\textbf{ImageNet accuracy of pre-training} 
Since our SupMAE is pre-trained in a supervised fashion on ImageNet, the pre-trained model can directly perform inference on the ImageNet dataset. We show the resulting performance in Table~\ref{tab:pt_acc}. The pre-trained (PT) and fine-tuned (FT) models are from Table~\ref{tab:main_results}. 
For the PT model, we test with two different patch ratios. When we keep the same ratio as for pre-training, \ie, 25\%, we achieve 66.0\% top1 accuracy. 
Interestingly, if we use all patches for inference, we can achieve an even higher accuracy of 68.9\%.
This shows that the pre-trained SupMAE model can achieve some level of accuracy even without fine-tuning.
It is worth noting that we do not expect the PT model to perform as well as the FT model. 
Table~\ref{tab:main_results} demonstrates that the pre-trained SupMAE model can provide a good initial representation point for other tasks.

\begin{table}[t]
\caption{\textbf{ImageNet-1K accuracy of pre-training} 
We report the ImageNet accuracy of pre-trained (PT) and fine-tuned (FT) SupMAE model on ImageNet-1K. All models are from Table~\ref{tab:main_results}.
(25\%) represents using 25\% patches for inference. 
Results show that even without fine-tuning, the pre-trained SupMAE model can achieve some level of accuracy on ImageNet-1K.
}
\label{tab:pt_acc}
\small
\centering
\begin{tabular}{lccc}
\toprule
          & PT (25\%) & PT (100\%) & FT \\
\midrule
Top1 acc. & 66.0          & 68.9          & 83.6 \\
\bottomrule
\end{tabular}
\end{table}

\begin{figure}[t]

\begin{center}
\centerline{\includegraphics[width=0.9\columnwidth]{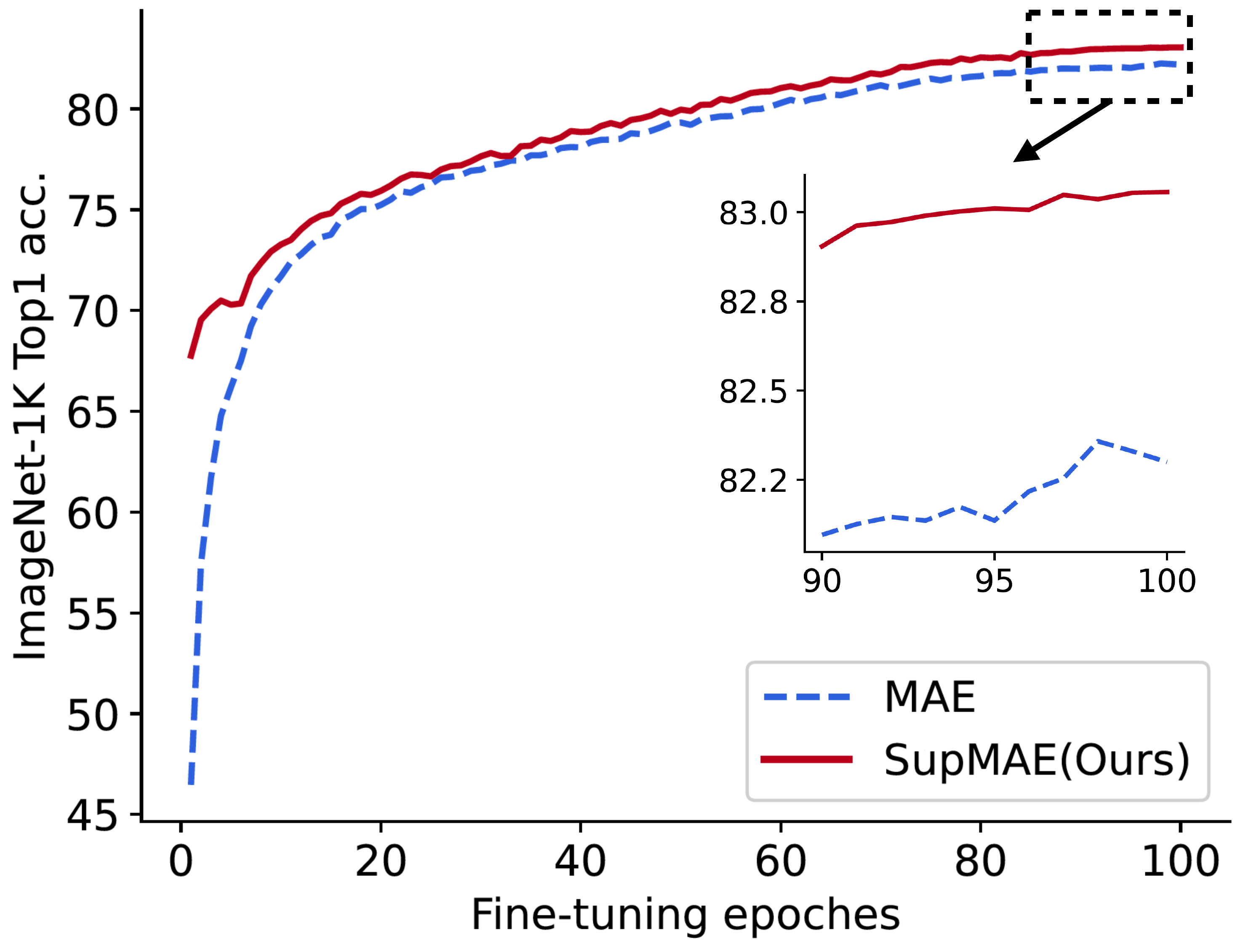}}
\caption{Comparison between MAE and SupMAE when fine-tuned for 100 epochs on ImageNet-1K. 
The model architecture is ViT-B/16.
Both MAE and SupMAE are pre-trained for 200 epochs.
Our SupMAE brings a much better initialization point than its MAE counterpart.
}
\label{fig:ft_acc}
\end{center}
\end{figure}

\textbf{ImageNet accuracy curve during fine-tuning} 
We further report the per epoch ImageNet accuracy during fine-tuning in Figure~\ref{fig:ft_acc}. 
MAE and SupMAE use ViT-B/16, and the models are pre-trained for 200 epochs.
Following the fine-tuning recipe in ~\citet{mae}, the pre-trained models are further fine-tuned on ImageNet for 100 epochs.
We can see that SupMAE can achieve very high accuracy even at the first epoch.
This is because our SupMAE can learn a good global feature via the supervised branch, which brings a better initialization point for fine-tuning.
Thanks to the good initialization, SupMAE can significantly improve over MAE when we only have a few samples in downstream datasets (See few-shot learning experiments in Table~\ref{tab:few-shot}).

\textbf{Transfer to SimMIM} Our introduced supervised branch is also supposed to be compatible with other self-supervised MIM methods.
To further validate the transferability, we extend the SimMIM~\cite{xie2021simmim} into a fully-supervised setting, like we do to MAE.
SimMIM predicts raw pixel values of the randomly masked patches by a lightweight one-layer head and performs learning using a simple $L_{1}$ loss.
The major difference between MAE and SimMIM is the input of the encoder.
SimMIM uses the corrupted image, where the masked patches are replaced by \texttt{mask} tokens (but it is still a complete image), while MAE only uses the visible patches as input.
When we extend the SiMMIM, we only use the visible patch features to generate global features, the same as SupMAE.

We validate our approach using the Swin-Base~\cite{swin} model to show that the supervised branch is also compatible with the hierarchical transformer. 
We set the loss ratio of the supervised classification branch as 0.01, the same as SupMAE.
All other training hyperparameters are the same as SimMIM.
We pre-train the models for 100 epochs and then fine-tune them for another 100 epochs.
The input resolutions are set as 192$\times$192.
The supervised branch can bring an additional +0.2\% improvement over the vanilla SimMIM.
It is worth noting that we do not do any hyperparameters tuning over SimMIM; more improvement can be expected if more careful tuning is done.

\begin{table}[t]
\caption{\textbf{Integrating the supervised branch into SimMIM} The model is Swin-Base. The pre-training and fine-tuning resolutions are all set to 192$\times$192. 
Results show that the supervised branch is also compatible with other MIM frameworks.
}
\label{tab:simmim}
\small
\centering

\begin{tabular}{lcc}
\toprule
method    & SimMIM & SimMIM w/ sup. \\
\midrule
Top1 acc. &   82.8     &   \textbf{83.0}          \\
\bottomrule
\end{tabular}
\end{table}

\section{Conclusion}
\label{sec:conclusion}

In this paper, we propose Supervised Masked Autoencoders (SupMAE), a \emph{fully-supervised} extension of MAE obtained by adding a supervised classification branch.
The supervised pre-training enables MAE to learn global features from golden labels effectively.
Unlike the standard supervised pre-training where all image patches are
used, the proposed SupMAE only exploits a visible subset of image patches for classification.
Through thorough experiments, we demonstrate that SupMAE is more training efficient and learns more robust and transferable features.

SupMAE is a hybrid pre-training method with external label supervision and self-provided supervision.
We believe that our method can shed light on future research on pre-training with multiple objectives.

\section{Acknowledgment}
\label{sec:acknowledgment}

This work was supported in part by NSF CCF Grant No. 2107085 ,ONR Minerva program, iMAGiNE - the Intelligent Machine Engineering Consortium at UT Austin, and a UT Cockrell School of Engineering Doctoral Fellowship.

\appendix
\section{Appendix}

\subsection{Implementation details on ImageNet}
\label{appendix:impl_details}

We follow the hyperparameters in MAE~\cite{mae}.

\paragraph{Pre-training.} The pre-training setting is in Table~\ref{tab:pretrain}. We train our SupMAE for 400 epochs with 20 epochs warm-up. The models are trained in a distributed fashion on 16 GPUs with a total batch size of 4096 (256 for each GPU). Like in MAE, we use the linear \textit{lr} scaling rule: \textit{lr} = \textit{base\_lr}$\times$batchsize / 256.

\begin{table}[h]
\tablestyle{6pt}{1.02}
\caption{\textbf{Pre-training hyperparameters.}}
\vspace{1em}
\begin{tabular}{ll}
\toprule
config & value \\
\midrule
optimizer & AdamW \\ 
base learning rate & 1.5e-4 \\
weight decay & 0.05 \\
optimizer momentum & $\beta_1, \beta_2{=}0.9, 0.95$ \\ 
batch size & 4096 \\
GPU number & 16 \\
learning rate schedule & cosine decay \\ 
pre-training epochs & 400 \\
warmup epochs  & 20 \\ 
augmentation & RandomResizedCrop \\
\bottomrule
\end{tabular}
\label{tab:pretrain}
\end{table}

\paragraph{End-to-end fine-tuning.} The fine-tuning setting is in Table~\ref{tab:finetune}. Like in MAE, we use the global pooling feature rather than \texttt{class} token during fine-tuning. The models are fine-tuned on 8 GPUs with a total batch size of 1024 (128 for each GPU). Like in the pre-training, we use the linear \textit{lr} scaling rule: \textit{lr} = \textit{base\_lr}$\times$batchsize / 256.

\begin{table}[t!]
\tablestyle{6pt}{1.02}
\caption{\textbf{End-to-end fine-tuning hyperparameters.}}
\vspace{1em}
\begin{tabular}{ll}
\toprule
config & value \\
\midrule
optimizer & AdamW \\
base learning rate & 1e-3 \\
weight decay & 0.05 \\
optimizer momentum & $\beta_1, \beta_2{=}0.9, 0.999$ \\
layer-wise lr decay & 0.65 \\ 
batch size & 1024 \\
learning rate schedule & cosine decay \\
warmup epochs & 5 \\
training epochs & 100 \\
augmentation & RandAug (9, 0.5) \\ 
label smoothing & 0.1 \\ 
reprob & 0.25 \\
mixup & 0.8 \\ 
cutmix & 1.0 \\ 
drop path & 0.1 \\ 
\bottomrule
\end{tabular}
\label{tab:finetune}
\end{table}

\paragraph{Linear probing.} The linear probing setting is in Table~\ref{tab:linearprob}. Like MAE, we use the \texttt{class} token during linear probing. The models are trained on 8 GPUs with a total batch size of 8192 (1024 for each GPU) with a LARS optimizer. A larger base \textit{lr} is used in linear probing.

\begin{table}[t!]
\tablestyle{6pt}{1.02}
\caption{\textbf{Linear probing hyperparameters.}}
\label{tab:linearprob}
\vspace{1em}
\begin{tabular}{ll}
\toprule
config & value \\
\midrule
optimizer & LARS \\ 
base learning rate & 0.1 \\
weight decay & 0 \\
optimizer momentum & 0.9 \\
batch size & 8192 \\
learning rate schedule & cosine decay \\
warmup epochs & 10 \\
training epochs & 90 \\
augmentation & RandomResizedCrop \\
\bottomrule
\end{tabular}
\end{table}


\subsection{Implementation details of few-shot learning}

We follow the few-shot transfer learning toolkit in ELEVATER~\cite{elevater}\footnote{https://github.com/Computer-Vision-in-the-Wild/Elevater\_Toolkit\_IC}.

\textbf{Datasets}

As summarized in Table~\ref{tab:downstream_ic_dataset}, ELEVATER~\cite{elevater} contains 20 image classification datasets, spans from common objects~\cite{krizhevsky2009learning,everingham2010pascal}, numerical digits~\cite{deng2012mnist}, aircraft images~\cite{maji2013fine}, satellite images~\cite{helber2019eurosat} and so on. 
We use the data samples defined in the ELEVATER toolkit when we perform few-shot learning.

\begin{table*}[t!]
  \centering
  \caption{\textbf{Statistics of 20 image classification datasets in ELEVATER~\cite{elevater}}.}
\label{tab:downstream_ic_dataset}
    \setlength{\tabcolsep}{2.2pt}
  \scalebox{0.8}{
\begin{tabular}{c | c c c c c c } 
 \toprule
 Dataset & \#Concepts & Train size & Test size & Evaluation metric & Source   \\ 
 \midrule

Hateful Memes~\cite{kiela2020hateful} & 2 & 8,500 & 500 & ROC AUC & Facebook \\
PatchCamelyon~\cite{veeling2018rotation}  & 2 & 262,144 & 32,768 & Accuracy & Tensorflow \\
Rendered-SST2~\cite{radford2021learning} & 2 & 6,920 & 1,821 & Accuracy & OpenAI \\
KITTI Distance~\cite{fritsch2013new} & 4 & 6,347 & 711 & Accuracy & KITTI website \\
FER 2013~\cite{fer2013} & 7 & 28,709 & 3,589 & Accuracy & Kaggle fer2013 \\
CIFAR-10~\cite{krizhevsky2009learning} & 10 & 50,000 & 10,000 & Accuracy & Tensorflow \\
EuroSAT~\cite{helber2019eurosat} & 10 & 5,000 & 5,000 & Accuracy & Tensorflow \\
MNIST~\cite{deng2012mnist} & 10 & 60,000 & 10,000 & Accuracy & Tensorflow \\
VOC 2007 Classification~\cite{everingham2010pascal} & 20 & 2,501 & 4,952 & 11-point mAP & VOC 2007 \\
Oxford-IIIT Pets~\cite{parkhi2012cats} & 37 & 3,680 & 3,669 & Mean-per-class & Oxford-IIIT Pets \\
GTSRB~\cite{stallkamp2011german} & 43 & 26,640 & 12,630 & Accuracy & GTSRB website \\
Resisc-45~\cite{cheng2017remote} & 45 & 3,150 & 25,200 & Accuracy & Tensorflow \\
Describable Textures~\cite{cimpoi2014describing} & 47 & 1,880 & 1,880 & Accuracy & DTD website \\
CIFAR-100~\cite{krizhevsky2009learning} & 100 & 50,000 & 10,000 & Accuracy & Tensorflow \\
FGVC Aircraft (variants)~\cite{maji2013fine} & 100 & 3,334 & 3,333 & Mean-per-class & FGVC website \\
Food-101~\cite{bossard2014food} & 101 & 75,750 & 25,250 & Accuracy & Tensorflow \\
Caltech-101~\cite{fei2004learning} & 102 & 3,060 & 6,084 & Mean-per-class & Tensorflow \\
Oxford Flowers 102~\cite{nilsback2008automated} & 102 & 1,020 & 6,149 & Mean-per-class & Tensorflow \\
Stanford Cars~\cite{krause20133d} & 196 & 8,144 & 8,041 & Accuracy & Stanford Cars \\
Country-211~\cite{radford2021learning} & 211 & 31,650 & 21,100 & Accuracy & OpenAI \\

\midrule
Total & 1151 & 638429 & 192677 & -- & -- &    \\ 

\bottomrule
\end{tabular}
}
\end{table*}

\begin{table*}[t!]
    \centering
    \caption{\textbf{Breakdown results of few-shot learning on 20 classification datasets.}}
    \label{tab:experiment_breakdown}
    \footnotesize
    \setlength{\tabcolsep}{2.2pt}
    \scalebox{0.8}{
    \begin{tabular}{cc|c|cccccccccccccccccccc}
        \toprule
        Ckpt. & Shot & Score & \rotatebox[origin=l]{90}{Caltech101} & \rotatebox[origin=l]{90}{CIFAR10} & \rotatebox[origin=l]{90}{CIFAR100} & \rotatebox[origin=l]{90}{Country211} & \rotatebox[origin=l]{90}{DTD} & \rotatebox[origin=l]{90}{EuroSat} & \rotatebox[origin=l]{90}{FER2013} & \rotatebox[origin=l]{90}{FGVCAircraft} & \rotatebox[origin=l]{90}{Food101} & \rotatebox[origin=l]{90}{GTSRB} & \rotatebox[origin=l]{90}{HatefulMemes} & \rotatebox[origin=l]{90}{KittiDistance} & \rotatebox[origin=l]{90}{MNIST} & \rotatebox[origin=l]{90}{Flowers102} & \rotatebox[origin=l]{90}{OxfordPets} & \rotatebox[origin=l]{90}{PatchCamelyon} & \rotatebox[origin=l]{90}{SST2} & \rotatebox[origin=l]{90}{RESISC45} & \rotatebox[origin=l]{90}{StanfordCars} & \rotatebox[origin=l]{90}{VOC2007} \\
        \midrule
        \multicolumn{23}{c}{ \bf{Linear Probing} } \\
\multirow{3}{*}{MAE} & 5 & 33.4 & 59.0 & 34.0 & 21.2 & 2.8 & 35.0 & 64.4 & 21.3 & 7.0 & 7.7 & 17.5 & 51.4 & 46.1 & 63.4 & 50.9 & 17.2 & 54.9 & 50.1 & 38.9 & 6.3 & 18.3 \\
 & 20 & 48.0 & 85.5 & 44.9 & 43.5 & 4.4 & 58.3 & 74.1 & 23.5 & 29.9 & 30.4 & 41.1 & 51.7 & 49.8 & 52.9 & 71.9 & 60.0 & 52.7 & \textbf{53.2} & 67.4 & 25.5 & 39.9 \\
 & 50 & 58.3 & 88.7 & 67.3 & 53.3 & 6.9 & 66.0 & 86.4 & 27.1 & 39.2 & 42.8 & 57.0 & 50.8 & 54.0 & 81.5 & 71.9 & 76.5 & 69.4 & 51.6 & 78.6 & 36.7 & 59.2 \\
\multirow{3}{*}{MoCo-v3} & 5 & 50.2 & 80.8 & 78.5 & 60.5 & 4.8 & 57.1 & 77.1 & 20.5 & 11.8 & 36.6 & 31.4 & 50.7 & 46.7 & 64.1 & 79.5 & 76.2 & 54.7 & 50.0 & 61.1 & 13.4 & 47.9 \\
 & 20 & 62.0 & 91.3 & 67.7 & 75.5 & 7.6 & 66.3 & 84.8 & 30.9 & 38.2 & 59.3 & 53.9 & \textbf{53.5} & 48.5 & 81.8 & 89.5 & 86.4 & 52.1 & 51.6 & 77.3 & 49.5 & 74.2 \\
 & 50 & \textbf{69.7} & \textbf{92.1} & \textbf{93.6} & \textbf{79.0} & \textbf{10.3} & \textbf{73.4} & \textbf{92.3} & \textbf{40.2} & \textbf{48.0} & \textbf{66.8} & \textbf{66.7} & 50.3 & 60.5 & 88.3 & 89.5 & \textbf{90.2} & 75.1 & 51.3 & 84.1 & \textbf{63.1} & \textbf{79.2} \\
\multirow{3}{*}{SupMAE (Ours)} & 5 & 48.0 & 73.9 & 55.1 & 38.5 & 4.4  & 44.4 & 77.0 & 19.2 & 16.2 & 33.8 & 31.1 & 47.2 & 42.8 & 73.3 & 82.7 & 72.0 & 66.8 & 50.4 & 61.9 & 16.7 & 52.2 \\
 & 20 & 60.9 & 87.6 & 70.6 & 54.5 & 7.3  & 61.7 & 89.7 & 23.8 & 33.2 & 53.6 & 55.9 & 50.0 & 57.2 & 87.6 & \textbf{90.5} & 85.4 & 76.1 & 51.1 & 76.2 & 36.2 & 69.3 \\
 & 50 & 66.7 & 88.5 & 80.1 & 65.6 & 10.0 & 68.0 & 92.1 & 34.9 & 40.3 & 62.3 & 64.2 & 50.7 & \textbf{61.6} & \textbf{94.3} & \textbf{90.5} & 88.7 & \textbf{80.4} & 52.5 & \textbf{84.4} & 48.3 & 76.2 \\
        \midrule
        \multicolumn{23}{c}{\bf{Fine-tuning}} \\
\multirow{3}{*}{MAE} & 5 & 36.1 & 70.8 & 34.4 & 13.1 & 2.1 & 41.4 & 64.1 & 20.8 & 8.2 & 13.3 & 14.8 & 49.6 & 38.0 & 46.8 & 68.8 & 37.8 & 53.3 & 50.9 & 50.4 & 6.0 & 37.4 \\
 & 20 & 54.1 & 91.0 & 50.1 & 40.4 & 3.6 & 59.7 & 79.5 & 22.6 & 32.5 & 22.4 & 62.2 & \textbf{54.8} & 46.0 & 90.9 & 81.6 & 78.0 & 67.7 & 51.7 & 65.5 & 21.6 & 60.8 \\
 & 50 & 65.9 & 92.9 & 71.5 & 54.7 & 4.9 & 66.2 & 87.8 & 34.1 & 42.8 & 51.9 & 95.6 & 51.3 & 50.1 & 96.1 & 81.6 & 84.8 & 77.3 & \textbf{52.4} & 85.2 & 68.0 & 68.0 \\
\multirow{3}{*}{MoCo-v3} & 5 & 39.3 & 73.7 & 70.3 & 17.4 & 2.3 & 45.6 & 60.0 & 13.5 & 7.2 & 27.6 & 16.5 & 50.8 & 43.5 & 18.1 & 65.7 & 77.1 & 50.9 & 50.7 & 58.2 & 11.2 & 25.7 \\
 & 20 & 58.8 & 91.9 & 58.4 & 59.2 & 5.0 & 63.4 & 69.7 & 19.8 & 47.4 & 55.5 & 86.7 & 53.5 & 48.5 & 53.4 & 85.8 & 87.4 & 51.5 & 51.4 & 78.5 & 49.2 & 59.2 \\
 & 50 & 70.3 & 92.8 & \textbf{89.1} & \textbf{77.5} & \textbf{6.9} & \textbf{71.3} & 92.6 & 31.0 & 53.4 & 63.2 & 96.5 & 50.9 & \textbf{57.3} & 94.3 & 85.8 & 90.2 & 74.2 & 50.4 & 87.3 & 66.2 & \textbf{75.7} \\
\multirow{3}{*}{SupMAE (Ours)} & 5 & 46.8 & 75.4 & 48.4 & 27.7 & 2.9 & 43.9 & 74.3 & 19.4 & 18.4 & 30.1 & 36.8 & 50.5 & 37.4 & 79.4 & 84.3 & 69.6 & 57.0 & 51.2 & 61.4 & 13.5 & 53.6 \\
 & 20 & 64.6 & 92.9 & 64.4 & 60.7 & 4.6 & 61.9 & 86.7 & 27.2 & 44.5 & 56.3 & 90.6 & 52.0 & 51.1 & 87.9 & 90.2 & 88.3 & 72.9 & 51.1 & 81.4 & 59.1 & 68.4 \\
 & 50 & \textbf{71.7} & \textbf{94.1} & 84.0 & 74.0 & 6.6 & 69.3 & \textbf{93.3} & \textbf{36.0} & \textbf{58.4} & \textbf{68.7} & \textbf{97.0} & 51.0 & \textbf{57.3} & \textbf{96.8} & \textbf{90.2} & \textbf{91.3} & \textbf{78.9} & 52.2 & \textbf{88.0} & \textbf{77.7} & 69.6 \\
        \bottomrule
    \end{tabular}
    }
\end{table*}

\textbf{Automatic hyperparameter tuning}
We also follow the automatic hyperparameter tuning strategy in the ELEVATER toolkit.
We split its training set into training (80\%) and validation (20\%) sets for each setting.
We will ensure that each category has at least one training sample.
For each setting, we first grid search the optimal learning rate $\eta$ and weight decay $\alpha$. 
In the hyper-parameter search phase, each configuration ($\eta$, $\alpha$) will run for 10 epochs.
After obtaining the optimal ($\eta$, $\alpha$), the final model will be trained for 50 epochs and tested on the testing set.

\textbf{Results breakdown}
We further report the results of each dataset and setting in Table~\ref{tab:experiment_breakdown}. The MAE and MoCo-v3 results are from ~\citet{elevater}. 
We note that our SupMAE model uses global pooling to extract global features in linear probing rather than the \texttt{class} token used by MAE and MoCO-v3.
This is because our SupMAE model is pre-trained with global pool mode.
For fine-tuning, all methods use global pooling.
Our SupMAE outperforms its MAE counterparts by a large margin, \eg, +14.5\% average score for 5-shot learning probing.
SupMAE shows more advantage under end-to-end fine-tuning setting, achieving 13 best accuracies among 20 datasets for 50-shot fine-tuning.

\bibliography{aaai24}

\begin{thebibliography}{62}
\providecommand{\natexlab}[1]{#1}

\bibitem[{Assran et~al.(2022)Assran, Caron, Misra, Bojanowski, Bordes, Vincent, Joulin, Rabbat, and Ballas}]{assran2022msn}
Assran, M.; Caron, M.; Misra, I.; Bojanowski, P.; Bordes, F.; Vincent, P.; Joulin, A.; Rabbat, M.; and Ballas, N. 2022.
\newblock Masked Siamese Networks for Label-Efficient Learning.
\newblock \emph{arXiv preprint arXiv:2204.07141}.

\bibitem[{Baevski et~al.(2022)Baevski, Hsu, Xu, Babu, Gu, and Auli}]{baevski2022data2vec}
Baevski, A.; Hsu, W.-N.; Xu, Q.; Babu, A.; Gu, J.; and Auli, M. 2022.
\newblock Data2vec: A general framework for self-supervised learning in speech, vision and language.
\newblock \emph{arXiv preprint arXiv:2202.03555}.

\bibitem[{Bao, Dong, and Wei(2021)}]{beit}
Bao, H.; Dong, L.; and Wei, F. 2021.
\newblock Beit: Bert pre-training of image transformers.
\newblock \emph{arXiv preprint arXiv:2106.08254}.

\bibitem[{Bossard, Guillaumin, and Gool(2014)}]{bossard2014food}
Bossard, L.; Guillaumin, M.; and Gool, L.~V. 2014.
\newblock Food-101--mining discriminative components with random forests.
\newblock In \emph{ECCV}.

\bibitem[{Caron et~al.(2021)Caron, Touvron, Misra, J{\'e}gou, Mairal, Bojanowski, and Joulin}]{dino}
Caron, M.; Touvron, H.; Misra, I.; J{\'e}gou, H.; Mairal, J.; Bojanowski, P.; and Joulin, A. 2021.
\newblock Emerging properties in self-supervised vision transformers.
\newblock In \emph{Proceedings of the IEEE/CVF International Conference on Computer Vision}, 9650--9660.

\bibitem[{Chen et~al.(2020)Chen, Radford, Child, Wu, Jun, Luan, and Sutskever}]{igpt}
Chen, M.; Radford, A.; Child, R.; Wu, J.; Jun, H.; Luan, D.; and Sutskever, I. 2020.
\newblock Generative pretraining from pixels.
\newblock In \emph{International conference on machine learning}, 1691--1703. PMLR.

\bibitem[{Chen*, Xie*, and He(2021)}]{mocov3}
Chen*, X.; Xie*, S.; and He, K. 2021.
\newblock An Empirical Study of Training Self-Supervised Vision Transformers.
\newblock \emph{arXiv preprint arXiv:2104.02057}.

\bibitem[{Cheng, Han, and Lu(2017)}]{cheng2017remote}
Cheng, G.; Han, J.; and Lu, X. 2017.
\newblock Remote sensing image scene classification: Benchmark and state of the art.
\newblock \emph{Proceedings of the IEEE}.

\bibitem[{Cimpoi et~al.(2014)Cimpoi, Maji, Kokkinos, Mohamed, and Vedaldi}]{cimpoi2014describing}
Cimpoi, M.; Maji, S.; Kokkinos, I.; Mohamed, S.; and Vedaldi, A. 2014.
\newblock Describing textures in the wild.
\newblock In \emph{CVPR}.

\bibitem[{Deng et~al.(2009)Deng, Dong, Socher, Li, Li, and Fei-Fei}]{imagenet}
Deng, J.; Dong, W.; Socher, R.; Li, L.-J.; Li, K.; and Fei-Fei, L. 2009.
\newblock Imagenet: A large-scale hierarchical image database.
\newblock In \emph{2009 IEEE conference on computer vision and pattern recognition}, 248--255. Ieee.

\bibitem[{Deng(2012)}]{deng2012mnist}
Deng, L. 2012.
\newblock The {MNIST} database of handwritten digit images for machine learning research.
\newblock \emph{IEEE signal processing magazine}.

\bibitem[{Devlin et~al.(2018)Devlin, Chang, Lee, and Toutanova}]{bert}
Devlin, J.; Chang, M.-W.; Lee, K.; and Toutanova, K. 2018.
\newblock Bert: Pre-training of deep bidirectional transformers for language understanding.
\newblock \emph{arXiv preprint arXiv:1810.04805}.

\bibitem[{Dong et~al.(2021)Dong, Bao, Zhang, Chen, Zhang, Yuan, Chen, Wen, and Yu}]{peco}
Dong, X.; Bao, J.; Zhang, T.; Chen, D.; Zhang, W.; Yuan, L.; Chen, D.; Wen, F.; and Yu, N. 2021.
\newblock Peco: Perceptual codebook for bert pre-training of vision transformers.
\newblock \emph{arXiv preprint arXiv:2111.12710}.

\bibitem[{Dosovitskiy et~al.(2020)Dosovitskiy, Beyer, Kolesnikov, Weissenborn, Zhai, Unterthiner, Dehghani, Minderer, Heigold, Gelly et~al.}]{vit}
Dosovitskiy, A.; Beyer, L.; Kolesnikov, A.; Weissenborn, D.; Zhai, X.; Unterthiner, T.; Dehghani, M.; Minderer, M.; Heigold, G.; Gelly, S.; et~al. 2020.
\newblock An image is worth 16x16 words: Transformers for image recognition at scale.
\newblock \emph{arXiv preprint arXiv:2010.11929}.

\bibitem[{Everingham et~al.(2010)Everingham, Van~Gool, Williams, Winn, and Zisserman}]{everingham2010pascal}
Everingham, M.; Van~Gool, L.; Williams, C.~K.; Winn, J.; and Zisserman, A. 2010.
\newblock The pascal visual object classes ({VOC}) challenge.
\newblock \emph{IJCV}.

\bibitem[{Fei-Fei, Fergus, and Perona(2004)}]{fei2004learning}
Fei-Fei, L.; Fergus, R.; and Perona, P. 2004.
\newblock Learning generative visual models from few training examples: An incremental {B}ayesian approach tested on 101 object categories.
\newblock In \emph{CVPR workshop}.

\bibitem[{Fritsch, Kuehnl, and Geiger(2013)}]{fritsch2013new}
Fritsch, J.; Kuehnl, T.; and Geiger, A. 2013.
\newblock A new performance measure and evaluation benchmark for road detection algorithms.
\newblock In \emph{ITSC}. IEEE.

\bibitem[{Girshick et~al.(2014)Girshick, Donahue, Darrell, and Malik}]{rcnn}
Girshick, R.; Donahue, J.; Darrell, T.; and Malik, J. 2014.
\newblock Rich feature hierarchies for accurate object detection and semantic segmentation.
\newblock In \emph{Proceedings of the IEEE conference on computer vision and pattern recognition}, 580--587.

\bibitem[{He et~al.(2021)He, Chen, Xie, Li, Doll{\'a}r, and Girshick}]{mae}
He, K.; Chen, X.; Xie, S.; Li, Y.; Doll{\'a}r, P.; and Girshick, R. 2021.
\newblock Masked autoencoders are scalable vision learners.
\newblock \emph{arXiv preprint arXiv:2111.06377}.

\bibitem[{He et~al.(2017)He, Gkioxari, Doll{\'a}r, and Girshick}]{he2017maskrcnn}
He, K.; Gkioxari, G.; Doll{\'a}r, P.; and Girshick, R. 2017.
\newblock Mask r-cnn.
\newblock In \emph{Proceedings of the IEEE international conference on computer vision}, 2961--2969.

\bibitem[{He et~al.(2016)He, Zhang, Ren, and Sun}]{resnet}
He, K.; Zhang, X.; Ren, S.; and Sun, J. 2016.
\newblock Deep residual learning for image recognition.
\newblock In \emph{Proceedings of the IEEE conference on computer vision and pattern recognition}, 770--778.

\bibitem[{Helber et~al.(2019)Helber, Bischke, Dengel, and Borth}]{helber2019eurosat}
Helber, P.; Bischke, B.; Dengel, A.; and Borth, D. 2019.
\newblock Euro{S}at: A novel dataset and deep learning benchmark for land use and land cover classification.
\newblock \emph{IEEE Journal of Selected Topics in Applied Earth Observations and Remote Sensing}.

\bibitem[{Hendrycks et~al.(2021{\natexlab{a}})Hendrycks, Basart, Mu, Kadavath, Wang, Dorundo, Desai, Zhu, Parajuli, Guo et~al.}]{imagenet-rendition}
Hendrycks, D.; Basart, S.; Mu, N.; Kadavath, S.; Wang, F.; Dorundo, E.; Desai, R.; Zhu, T.; Parajuli, S.; Guo, M.; et~al. 2021{\natexlab{a}}.
\newblock The many faces of robustness: A critical analysis of out-of-distribution generalization.
\newblock In \emph{Proceedings of the IEEE/CVF International Conference on Computer Vision}, 8340--8349.

\bibitem[{Hendrycks and Dietterich(2019)}]{imagenet-corruption}
Hendrycks, D.; and Dietterich, T. 2019.
\newblock Benchmarking neural network robustness to common corruptions and perturbations.
\newblock \emph{arXiv preprint arXiv:1903.12261}.

\bibitem[{Hendrycks et~al.(2021{\natexlab{b}})Hendrycks, Zhao, Basart, Steinhardt, and Song}]{imagenet-adversarial}
Hendrycks, D.; Zhao, K.; Basart, S.; Steinhardt, J.; and Song, D. 2021{\natexlab{b}}.
\newblock Natural adversarial examples.
\newblock In \emph{Proceedings of the IEEE/CVF Conference on Computer Vision and Pattern Recognition}, 15262--15271.

\bibitem[{Hinton et~al.(2015)Hinton, Vinyals, Dean et~al.}]{hinton2015distilling}
Hinton, G.; Vinyals, O.; Dean, J.; et~al. 2015.
\newblock Distilling the knowledge in a neural network.
\newblock \emph{arXiv preprint arXiv:1503.02531}, 2(7).

\bibitem[{Huang et~al.(2022)Huang, Jin, Lu, Hou, Cheng, Fu, Shen, and Feng}]{cmae}
Huang, Z.; Jin, X.; Lu, C.; Hou, Q.; Cheng, M.-M.; Fu, D.; Shen, X.; and Feng, J. 2022.
\newblock Contrastive Masked Autoencoders are Stronger Vision Learners.
\newblock \emph{arXiv preprint arXiv:2207.13532}.

\bibitem[{Ioffe and Szegedy(2015)}]{ioffe2015batchnorm}
Ioffe, S.; and Szegedy, C. 2015.
\newblock Batch normalization: Accelerating deep network training by reducing internal covariate shift.
\newblock In \emph{International conference on machine learning}, 448--456. PMLR.

\bibitem[{kaggle(2013)}]{fer2013}
kaggle. 2013.
\newblock {FER} 2013: Kaggle challenges in representation learning facial expression recognition.
\newblock \url{https://www.kaggle.com/}.

\bibitem[{Khosla et~al.(2020)Khosla, Teterwak, Wang, Sarna, Tian, Isola, Maschinot, Liu, and Krishnan}]{supcon}
Khosla, P.; Teterwak, P.; Wang, C.; Sarna, A.; Tian, Y.; Isola, P.; Maschinot, A.; Liu, C.; and Krishnan, D. 2020.
\newblock Supervised contrastive learning.
\newblock \emph{Advances in Neural Information Processing Systems}, 33: 18661--18673.

\bibitem[{Kiela et~al.(2020)Kiela, Firooz, Mohan, Goswami, Singh, Ringshia, and Testuggine}]{kiela2020hateful}
Kiela, D.; Firooz, H.; Mohan, A.; Goswami, V.; Singh, A.; Ringshia, P.; and Testuggine, D. 2020.
\newblock The hateful memes challenge: Detecting hate speech in multimodal memes.
\newblock \emph{NeurIPS}.

\bibitem[{Krause et~al.(2013)Krause, Stark, Deng, and Fei-Fei}]{krause20133d}
Krause, J.; Stark, M.; Deng, J.; and Fei-Fei, L. 2013.
\newblock 3d object representations for fine-grained categorization.
\newblock In \emph{ICCV workshops}.

\bibitem[{Krizhevsky, Hinton et~al.(2009)}]{krizhevsky2009learning}
Krizhevsky, A.; Hinton, G.; et~al. 2009.
\newblock Learning multiple layers of features from tiny images.

\bibitem[{Krizhevsky, Sutskever, and Hinton(2012)}]{alexnet}
Krizhevsky, A.; Sutskever, I.; and Hinton, G.~E. 2012.
\newblock Imagenet classification with deep convolutional neural networks.
\newblock \emph{Advances in neural information processing systems}, 25: 1097--1105.

\bibitem[{Le, Patterson, and White(2018)}]{le2018supervised}
Le, L.; Patterson, A.; and White, M. 2018.
\newblock Supervised autoencoders: Improving generalization performance with unsupervised regularizers.
\newblock \emph{Advances in neural information processing systems}, 31.

\bibitem[{Li et~al.(2022{\natexlab{a}})Li, Liu, Li, Zhang, Aneja, Yang, Jin, Lee, Hu, Liu et~al.}]{elevater}
Li, C.; Liu, H.; Li, L.~H.; Zhang, P.; Aneja, J.; Yang, J.; Jin, P.; Lee, Y.~J.; Hu, H.; Liu, Z.; et~al. 2022{\natexlab{a}}.
\newblock ELEVATER: A Benchmark and Toolkit for Evaluating Language-Augmented Visual Models.
\newblock \emph{arXiv preprint arXiv:2204.08790}.

\bibitem[{Li et~al.(2022{\natexlab{b}})Li, Wu, Wu, Zang, Wang, Shang, Sun, Li, Li et~al.}]{a2mim}
Li, S.; Wu, D.; Wu, F.; Zang, Z.; Wang, K.; Shang, L.; Sun, B.; Li, H.; Li, S.; et~al. 2022{\natexlab{b}}.
\newblock Architecture-Agnostic Masked Image Modeling--From ViT back to CNN.
\newblock \emph{arXiv preprint arXiv:2205.13943}.

\bibitem[{Liu et~al.(2021)Liu, Lin, Cao, Hu, Wei, Zhang, Lin, and Guo}]{swin}
Liu, Z.; Lin, Y.; Cao, Y.; Hu, H.; Wei, Y.; Zhang, Z.; Lin, S.; and Guo, B. 2021.
\newblock Swin transformer: Hierarchical vision transformer using shifted windows.
\newblock In \emph{Proceedings of the IEEE/CVF International Conference on Computer Vision}, 10012--10022.

\bibitem[{Long, Shelhamer, and Darrell(2015)}]{fcn}
Long, J.; Shelhamer, E.; and Darrell, T. 2015.
\newblock Fully convolutional networks for semantic segmentation.
\newblock In \emph{Proceedings of the IEEE conference on computer vision and pattern recognition}, 3431--3440.

\bibitem[{Maji et~al.(2013)Maji, Rahtu, Kannala, Blaschko, and Vedaldi}]{maji2013fine}
Maji, S.; Rahtu, E.; Kannala, J.; Blaschko, M.; and Vedaldi, A. 2013.
\newblock Fine-grained visual classification of aircraft.
\newblock \emph{arXiv preprint arXiv:1306.5151}.

\bibitem[{MMSegmentation(2020)}]{mmseg2020}
MMSegmentation, C. 2020.
\newblock {MMSegmentation}: OpenMMLab Semantic Segmentation Toolbox and Benchmark.
\newblock \url{https://github.com/open-mmlab/mmsegmentation}.

\bibitem[{Nilsback and Zisserman(2008)}]{nilsback2008automated}
Nilsback, M.-E.; and Zisserman, A. 2008.
\newblock Automated flower classification over a large number of classes.
\newblock In \emph{Indian Conference on Computer Vision, Graphics \& Image Processing}. IEEE.

\bibitem[{Parkhi et~al.(2012)Parkhi, Vedaldi, Zisserman, and Jawahar}]{parkhi2012cats}
Parkhi, O.~M.; Vedaldi, A.; Zisserman, A.; and Jawahar, C. 2012.
\newblock Cats and dogs.
\newblock In \emph{CVPR}.

\bibitem[{Radford et~al.(2021)Radford, Kim, Hallacy, Ramesh, Goh, Agarwal, Sastry, Askell, Mishkin, Clark et~al.}]{radford2021learning}
Radford, A.; Kim, J.~W.; Hallacy, C.; Ramesh, A.; Goh, G.; Agarwal, S.; Sastry, G.; Askell, A.; Mishkin, P.; Clark, J.; et~al. 2021.
\newblock Learning transferable visual models from natural language supervision.
\newblock In \emph{ICML}.

\bibitem[{Radford et~al.(2018)Radford, Narasimhan, Salimans, Sutskever et~al.}]{GPT}
Radford, A.; Narasimhan, K.; Salimans, T.; Sutskever, I.; et~al. 2018.
\newblock Improving language understanding by generative pre-training.

\bibitem[{Simonyan and Zisserman(2014)}]{vgg}
Simonyan, K.; and Zisserman, A. 2014.
\newblock Very deep convolutional networks for large-scale image recognition.
\newblock \emph{arXiv preprint arXiv:1409.1556}.

\bibitem[{Stallkamp et~al.(2011)Stallkamp, Schlipsing, Salmen, and Igel}]{stallkamp2011german}
Stallkamp, J.; Schlipsing, M.; Salmen, J.; and Igel, C. 2011.
\newblock The German traffic sign recognition benchmark: a multi-class classification competition.
\newblock In \emph{IJCNN}.

\bibitem[{Steiner et~al.(2021)Steiner, Kolesnikov, Zhai, Wightman, Uszkoreit, and Beyer}]{howtotrainvit}
Steiner, A.; Kolesnikov, A.; Zhai, X.; Wightman, R.; Uszkoreit, J.; and Beyer, L. 2021.
\newblock How to train your vit? data, augmentation, and regularization in vision transformers.
\newblock \emph{arXiv preprint arXiv:2106.10270}.

\bibitem[{Touvron et~al.(2021)Touvron, Cord, Douze, Massa, Sablayrolles, and J{\'e}gou}]{deit}
Touvron, H.; Cord, M.; Douze, M.; Massa, F.; Sablayrolles, A.; and J{\'e}gou, H. 2021.
\newblock Training data-efficient image transformers \& distillation through attention.
\newblock In \emph{International Conference on Machine Learning}, 10347--10357. PMLR.

\bibitem[{Touvron, Cord, and J{\'e}gou(2022)}]{touvron2022deit3}
Touvron, H.; Cord, M.; and J{\'e}gou, H. 2022.
\newblock DeiT III: Revenge of the ViT.
\newblock \emph{arXiv preprint arXiv:2204.07118}.

\bibitem[{Vaswani et~al.(2017)Vaswani, Shazeer, Parmar, Uszkoreit, Jones, Gomez, Kaiser, and Polosukhin}]{transformer}
Vaswani, A.; Shazeer, N.; Parmar, N.; Uszkoreit, J.; Jones, L.; Gomez, A.~N.; Kaiser, {\L}.; and Polosukhin, I. 2017.
\newblock Attention is all you need.
\newblock \emph{Advances in neural information processing systems}, 30.

\bibitem[{Veeling et~al.(2018)Veeling, Linmans, Winkens, Cohen, and Welling}]{veeling2018rotation}
Veeling, B.~S.; Linmans, J.; Winkens, J.; Cohen, T.; and Welling, M. 2018.
\newblock Rotation equivariant CNNs for digital pathology.
\newblock In \emph{MICCAI}.

\bibitem[{Wang et~al.(2017)Wang, Xiang, Cheng, and Yuille}]{wang2017normface}
Wang, F.; Xiang, X.; Cheng, J.; and Yuille, A.~L. 2017.
\newblock Normface: L2 hypersphere embedding for face verification.
\newblock In \emph{Proceedings of the 25th ACM international conference on Multimedia}, 1041--1049.

\bibitem[{Wang et~al.(2019)Wang, Ge, Lipton, and Xing}]{imagenet-sketch}
Wang, H.; Ge, S.; Lipton, Z.; and Xing, E.~P. 2019.
\newblock Learning robust global representations by penalizing local predictive power.
\newblock \emph{Advances in Neural Information Processing Systems}, 32.

\bibitem[{Wang et~al.(2022)Wang, Liang, Li, Ouyang, Zhang, and Shao}]{repre}
Wang, L.; Liang, F.; Li, Y.; Ouyang, W.; Zhang, H.; and Shao, J. 2022.
\newblock Repre: Improving self-supervised vision transformer with reconstructive pre-training.
\newblock \emph{arXiv preprint arXiv:2201.06857}.

\bibitem[{Wei et~al.(2021)Wei, Fan, Xie, Wu, Yuille, and Feichtenhofer}]{wei2021masked}
Wei, C.; Fan, H.; Xie, S.; Wu, C.-Y.; Yuille, A.; and Feichtenhofer, C. 2021.
\newblock Masked Feature Prediction for Self-Supervised Visual Pre-Training.
\newblock \emph{arXiv preprint arXiv:2112.09133}.

\bibitem[{Wu et~al.(2018)Wu, Xiong, Yu, and Lin}]{wu2018unsupervised}
Wu, Z.; Xiong, Y.; Yu, S.~X.; and Lin, D. 2018.
\newblock Unsupervised feature learning via non-parametric instance discrimination.
\newblock In \emph{Proceedings of the IEEE conference on computer vision and pattern recognition}, 3733--3742.

\bibitem[{Xiao et~al.(2018)Xiao, Liu, Zhou, Jiang, and Sun}]{upernet}
Xiao, T.; Liu, Y.; Zhou, B.; Jiang, Y.; and Sun, J. 2018.
\newblock Unified perceptual parsing for scene understanding.
\newblock In \emph{Proceedings of the European conference on computer vision (ECCV)}, 418--434.

\bibitem[{Xie et~al.(2021{\natexlab{a}})Xie, Lin, Yao, Zhang, Dai, Cao, and Hu}]{moby}
Xie, Z.; Lin, Y.; Yao, Z.; Zhang, Z.; Dai, Q.; Cao, Y.; and Hu, H. 2021{\natexlab{a}}.
\newblock Self-supervised learning with swin transformers.
\newblock \emph{arXiv preprint arXiv:2105.04553}.

\bibitem[{Xie et~al.(2021{\natexlab{b}})Xie, Zhang, Cao, Lin, Bao, Yao, Dai, and Hu}]{xie2021simmim}
Xie, Z.; Zhang, Z.; Cao, Y.; Lin, Y.; Bao, J.; Yao, Z.; Dai, Q.; and Hu, H. 2021{\natexlab{b}}.
\newblock Simmim: A simple framework for masked image modeling.
\newblock \emph{arXiv preprint arXiv:2111.09886}.

\bibitem[{Zhang, Lee, and Lee(2016)}]{zhang2016augmenting}
Zhang, Y.; Lee, K.; and Lee, H. 2016.
\newblock Augmenting supervised neural networks with unsupervised objectives for large-scale image classification.
\newblock In \emph{International conference on machine learning}, 612--621. PMLR.

\bibitem[{Zhou et~al.(2019)Zhou, Zhao, Puig, Xiao, Fidler, Barriuso, and Torralba}]{ade20k}
Zhou, B.; Zhao, H.; Puig, X.; Xiao, T.; Fidler, S.; Barriuso, A.; and Torralba, A. 2019.
\newblock Semantic understanding of scenes through the ade20k dataset.
\newblock \emph{International Journal of Computer Vision}, 127(3): 302--321.

\end{thebibliography}

\end{document}